%% file: main.tex
\documentclass[10pt, conference]{IEEEtran}
\IEEEoverridecommandlockouts

\usepackage{pgfplots}
\usepackage{pgfplotstable}
\usepgfplotslibrary{fillbetween}
\usepackage{cite}
\usepackage{comment}
\usepackage{dsfont}
\usepackage{amsmath,amssymb,amsfonts}
\usepackage{algorithmic}
\usepackage{graphicx}
\usepackage{subfigure}
\usepackage{caption}
\usepackage{textcomp}
\usepackage{xcolor}
\usepackage{mdframed}
\def\BibTeX{{\rm B\kern-.05em{\sc i\kern-.025em b}\kern-.08em
    T\kern-.1667em\lower.7ex\hbox{E}\kern-.125emX}}
\usepackage[english]{babel}
\usepackage{tabularx}
\usepackage{booktabs}
\usepackage[justification=centering]{caption}
\usepackage{amsmath}
\usepackage{amsthm}
\newtheorem{theorem}{Theorem}

\usepackage{amssymb}
\usepackage{algorithm}
\usepackage{algorithmic}
\usepackage[colorlinks=true, allcolors=blue]{hyperref}

\newcommand{\figsize}{1\columnwidth}

\usepackage{paralist} 
\setdefaultleftmargin{1em}{0em}{0em}{0em}{0em}{0em} 
\usepackage{enumitem}

\usepackage{mathtools}
\usepackage{bbm}
\usepackage{stackengine}
\makeatletter
\newcommand\underparen[1]{\@ifnextchar_{\uphelp{\uparen{#1}}}{\uparen{#1}}}
\makeatother
\def\uphelp#1_#2{\ensurestackMath{\stackunder[1pt]{#1}{\scriptstyle #2}}}
\newcommand\uparen[1]{\setbox0=\hbox{$#1$}\ensurestackMath{%
  \stackunder[0pt]{#1}{\rotatebox{90}{$\left(%
  \rule[\dimexpr-.5\wd0+\dp\strutbox-1.3pt]{0pt}{\wd0}\right.$}}%
}}
\usepackage[
  paper   = letterpaper,     
  top     = 0.75in,          
  bottom  = 1.00in,          
  left    = 0.69in,
  right   = 0.69in,
  includehead,   
  includefoot,   
  heightrounded  
]{geometry}
\usepackage{amssymb} 

\usepackage{xspace}

\newcommand{\keepcomment}{1} 

\usepackage[normalem]{ulem}

\ifnum\keepcomment=1
	\usepackage[colorinlistoftodos,textsize=scriptsize]{todonotes} 
    \newcommand{\stkout}[1]{\ifmmode\text{\sout{\ensuremath{#1}}}\else\sout{#1}\fi}
    
\else
	
	\usepackage[disable]{todonotes} 
\fi

\begin{document}

\title{Online Learning for Function Placement in Serverless Computing}

\author{
    \IEEEauthorblockN{Wei Huang}
    \IEEEauthorblockA{
        \textit{SAMOVAR, Télécom SudParis, IP-Paris} \\
        \textit{ 91120 Palaiseau, France }\\
        wei.huang@telecom-sudparis.eu
    }
    \and
    \IEEEauthorblockN{Richard Combes}
    \IEEEauthorblockA{
        \textit{L2S, CentraleSupelec, Paris-Saclay University} \\
        \textit{91190 Gif-sur-Yvette, France}\\
        richard.combes@centralesupelec.fr
    }
    \and
    \IEEEauthorblockN{Andrea Araldo}
    \IEEEauthorblockA{
        \textit{SAMOVAR, Télécom SudParis, IP-Paris} \\
        \textit{ 91120 Palaiseau, France }\\
        andrea.araldo@telecom-sudparis.eu
    }
    \and
    \IEEEauthorblockN{Hind Castel-Taleb}
    \IEEEauthorblockA{
        \textit{SAMOVAR, Télécom SudParis, IP-Paris} \\
        \textit{ 91120 Palaiseau, France }\\
        hind.castel@telecom-sudparis.eu
    }
    \and
    \IEEEauthorblockN{Badii Jouaber}
    \IEEEauthorblockA{
       \textit{SAMOVAR, Télécom SudParis, IP-Paris} \\
        \textit{ 91120 Palaiseau, France }\\
        badii.jouaber@telecom-sudparis.eu
    }
}

\maketitle

\input{abstract}

\input{introduction}
\input{model}
\input{algorithm}
\input{analysis}
\input{numerical_results}

\input{conclusion}
\input{acknowledgment}

\bibliographystyle{IEEEtran}
\bibliography{ref} 


\end{document}

%% file: abstract.tex
\begin{abstract}
We study the placement of virtual functions aimed at minimizing the cost. We propose a novel algorithm, using ideas based on multi-armed bandits. We prove that these algorithms learn the optimal placement policy rapidly, and their regret grows at a rate at most $O( N M \sqrt{T\ln T} )$ while respecting the feasibility constraints with high probability, where $T$ is total time slots, $M$ is the number of classes of function and $N$ is the number of computation nodes. We show through numerical experiments that the proposed algorithm both has good practical performance and modest computational complexity. We propose an acceleration technique that allows the algorithm to achieve good performance also in large networks where computational power is limited. Our experiments are fully reproducible, and the code is publicly available. 
\end{abstract}


%% file: introduction.tex
\section{Introduction}\label{section:introduction}

In modern networks, some nodes usually have a high computing power and can perform various advanced functions far beyond switching and routing. With the increasing adoption of cloud/edge computing, Virtual Function Placement (VFP) plays a key role in managing network resources flexibly and efficiently. We consider the VFP problem, where virtual functions (VFs) enter a physical network sequentially and must be placed at physical nodes, with various resources {available} (for instance, computation, memory, and power). VFP is an essential problem in the field of network virtualization. There are two main goals: (1) let the virtual networks operate without hindering each other and share the physical resources seamlessly, (2) keep the cost of operating the network low (for instance, keep power consumption to a minimum).

The problem of placing VNFs onto a physical network is called Virtual Network Embedding (VNE). VNE is often studied together with routing requests in a virtual network~\cite{sfc_placment_routing} and offloading computation tasks from user devices to the VFs~\cite{Grosu2020,sfc_placement_tpds} or placing virtual machines in the edge~\cite{Grosu2022}. This problem is usually a complicated, dynamic problem with various conflicting objectives. Furthermore, as in all networking problems, the decision-maker must consider various sources of randomness, such as traffic, the arrival of virtual functions, and the availability of resources. Ideally, the decision maker should be able to "learn" the optimal placement by trial and error using online learning mechanisms such as multi-armed bandits and reinforcement learning. This work focuses on online learning mechanisms for the VFP problem.

\paragraph*{Our Contribution}
Our main contribution is to propose a novel algorithm for the VFP problem. The algorithms leverage ideas from multi-armed bandits~\cite{slivkins2019introduction,kl_ucb}, adapted to a different setting where the policy is a set of probability distributions and the learner must respect probabilistic constraints. Those algorithms are simple to implement, have solid theoretical performance guarantees (they are not a heuristic), and learn the optimal policy quickly, especially in their accelerated version. Fast learning is necessary when the network conditions frequently change, so one must relaunch the learning process often to track and adapt to the changing network conditions. 

The rest of the paper is organized as follows. In Section~\ref{Related_work}, we review the relevant literature and discuss how our work differs from previous studies. Section~\ref{section:model} introduces our model for the virtual function placement problem (VFP). Our proposed algorithms are presented in Section~\ref{section:algorithm}, with their theoretical performance analyzed in Section~\ref{section:analysis}. We evaluate the practical performance of our methods through numerical experiments in Section~\ref{section:numerical}. Section~\ref{section:conclusion} concludes the paper.

\section{Related Work}
\label{Related_work}
Prior work has considered VFP in a variety of settings. \cite{dehury2019} considers two types of resources, and the objective is to maximize resource utilization (not minimizing cost subject to resource constraints). The proposed algorithm has polynomial complexity, but it is a heuristic algorithm. \cite{faas_placement} develops and evaluates a statistical machine learning-based adaptive function placement technique designed to optimize function execution performance while minimizing operational costs on serverless computing platforms. Both \cite{yan2020} and \cite{zhang2022} aim at maximizing the acceptance ratio and the revenue (similar to minimizing cost). They use a deep reinforcement learning approach using convolutional neural networks, which requires extensive training time, approximately several days, using GPUs. However, the above work cannot provide any theoretical guarantees. \cite{yao2021} aims to maximize the ratio between revenue and cost and considers only a single type of resource (CPU). They design algorithms assuming that a single virtual function does not use a large proportion of the available resources and base their approach on matrix perturbation theory. This assumption is critical for the results to work. \cite{fu2021} considers the more general problem of placement of virtual networks comprised of several virtual functions. Admission control applies, and the system may reject incoming virtual networks if no resources are available. \cite{elkael2022} considers a single type of resource (CPU) under constraints on the bandwidth of the links between the nodes. As in \cite{dehury2019}, the goal is to maximize resource utilization. The solution uses reinforcement learning, more precisely, Monte-Carlo Tree Search, which is computationally expensive. \cite{maleki2023} considers a more complex problem where the system relies on sensing to allocate functions to nodes, and the set of nodes at which a function can be placed depends on this sensing process (the critical parameter being the sensing range). 


%% file: model.tex
\section{Model and objectives}\label{section:model}
In this section, we present the model and objectives of our problem, outlining the system structure and performance criteria. We introduce the proposed model and key statistical assumptions in Section~\ref{subsection:model}. Then, we present practical assumptions to show the applicability of our model in Section~\ref{subsection:assumptions} and formulate the optimization problem under optimal static policy in Section~\ref{static}. Next, in Section~\ref{regret} we discuss the regret minimization approach. In Section~\ref{generalized}, we generalize the model to ensure feasibility in all scenarios.


\subsection{Model}\label{subsection:model}
{We consider a system composed of $N$ computation nodes on which $M$ classes of function can be run, under the Function as a Service (FaaS) paradigm.
Nodes are equipped with $K$ types of resources. Nodes are capacitated: node~$i\in [N]$ has~$\beta_{i,k}$ amount of resources of type~$k, \forall k\in [K]$. Such capacities are assumed to be known and constant.
Users send to the system functions to be executed. A central controller decides where to ``place'' each function, i.e., in which node each function should run. After the function is placed, the corresponding computation is executed, the function disappears from the system and the results are provided to the user.  Each function belongs to one of $M$ classes.
Functions enter the system over time, with random inter-arrival time. Let~$\Delta t>0$ denote the minimum inter-arrival time, which can be very small and which depends on the physical limitations of the system. For instance, functions might sequentially arrive to the system through a communication channel, requiring at least~$\Delta t$ for each transmission. We consider the system over a finite time horizon, which we discretize in~$T$ time slots, each of duration~$\Delta t$. 
At each time slot $t \in [T]$, either no function enters the system or a function of class $j(t) \in [M]$ enters the system. For simplicity of notation, we assume that $M$ is a fictitious ``null'' class: when $j(t)=M$, it means that no function has entered the system in~$t$.
}

{Each incoming function has specific requirements. In the simplest case, these requirements indicate directly the amount of resource of each time that is requested to run the function. However, requirements might be more complex, i.e., they could be expressed in terms of latency constraints: a user might require that the latency to execute the function must be below a certain threshold, with high probability (see~\cite[(12)]{elkael2023joint}).
In this case, the controller would need to calculate the amount of resources to be reserved to the function in order to match such latency constraints. In the most general case, the amount of required resources might depend to the node in which the function is placed. Indeed, if the function is placed in a certain node, the communication delay between such a node and the user has an impact on the amount of required computation resources: the larger the communication delays, the smaller the computation delay should be (in order to meet the latency requirements) and the larger the amount of computation resources that should be allocated to that function.
}

We assume that a function of class~$j\in [M]$ requires amount~$A_{i,j,k}$ of resource of type~$k\in [K]$ when it is placed in node~$i\in [N]$. For instance, a function might need to consume more resource when it is placed in a node far away from the end user. Indeed, in such a case, to prevent overall delay from exploding, elaboration time should be reduced via using more resources. We assume~$A_{i,j,k}$ is a random variable. Note that the realizations $A_{i,j,k}$ are observed by the controller after the functions arrive but before they are placed to the system for all $i \in [N]$ and $k \in [K]$.

%

%

{Upon the arrival of a function at time~$t$, t}he controller {decides node~$i(t) \in [N]$ in which the function will be} place{d. After the placement, the controller} observes {a} cost {
of placement, which is modelled as follows. We assume that placing a function of class~$j$ in node~$i$ has cost~$C_{i,j}$, which is an unknown random variable. The cost of placing the function at time slot~$t$ is thus a realization of such a random variable, which we denote as~$C_{i,j}(t)\in[0,1]$, where~$i=i(t)$ is the node in which the function is placed and~$j=j(t)$ is the class of function that is placed in time slot $t$.
The fictitious ``null'' function class~$M$ has always 0 cost and 0 resource requirements.
}
{The amount of resources of type $k$ available at node $i$ per time slot is assumed fixed and known and denoted by $\beta_{i,k}$.}

We make the following statistical assumptions:
\begin{enumerate}
    \item for all $i,j$, $(C_{i,j}(t))_{t \ge 0}$ is i.i.d. with mean $c_{i,j} \in [0,1]$ (the mean cost of placing a function of type $j$ at node $i$).
    \item  for all $i,j,k$, {sequence of resource requirements} $(A_{i,j,k}(t))_{t \ge 0}$ is i.i.d. with mean $a_{i,j,k} \in [0,1]$ (the average amount of resource $k$ necessary to place a function of class $j$ at node $i$).
    \item {function class} $(j(t))_{t\ge0}$ is i.i.d. with distribution $\lambda${$=(\lambda_j)_{j\in [M]}$} in the sense that $\mathbb{P}(j(t)=j)=\lambda_j$.
\end{enumerate}
 We emphasize that {$\left(A_{i,j,k}(t)\right)_{t>0}, \left(C_{i,j}(t)\right)_{t>0}$ and $\left(j(t)\right)_{t>0}$} are random processes initially unknown to the controller{. Therefore,} to select $i(t)$ at time $t$, the only information available is {previously observed realizations} $A_{i,j(s),k}(s)$ {and~$C_{i(s),j(s)}(s)$} for all $i \in [N]$ all $k \in [K]$ and all $s < t$. As a consequence, the decisions yielding a low cost are not known apriori and must be learned by trial and error, i.e., by trying different associations of nodes and functions.

{
Let~$I_j(t)$ be the indicator function equal to~$1$ if the function entering the system at time~$t$ is~$j$ and equal to~$0$ otherwise.
At time~$t$, the controller makes decision~$I_{i,j}(t)$: value~$I_{i,j}(t)=1$ means that if a function of class~$j$ enters the system, it will be placed in node~$i$, otherwise~$I_{i,j}(t)=0$. Obviously, only~$I_{i,j}(t)=1$ for~$j=j(t)$ takes effect (the controller places at time~$t$ only the function that actually arrives at time~$t$.}
The objective is to minimize the average cost under the constraint that{, on average, the resources in each node are not exceeded. The problem that the controller needs to solve is}



\begin{align}
    \label{eq:resource-object}
    \text{ min } & {1  \over T} \sum_{t=1}^T \sum_{j=1}^M \sum_{i=1}^N I_j(t) \cdot I_{i,j}(t) \cdot C_{i,j}(t) 
    \\
    \label{eq:resource-constraint}
    \text{ s.t. } & {1 \over T} \sum_{t=1}^T \sum_{j=1}^M I_{j}(t)  \cdot I_{i,j}(t) \cdot A_{i,j,k}(t) \le \beta_{i,k} \;\; \forall i,k
    \\
    & \sum_{i=1}^N I_{i,j}(t)=1, \forall j\in[M]
    \label{eq:resource-constraint_1}
\end{align}

\subsection{Assumptions}\label{subsection:assumptions}
Our model is generic, and in practice, the types of resources available at nodes may represent, for instance, computation (CPU) and memory (RAM), and the costs may represent, for example, power consumption or financial costs necessary to operate the nodes. Both costs incurred and resources consumed by each function are stochastic. The model captures the fluctuations occurring in real-world systems due to the randomness of traffic, noise, etc. For instance, {the resources consumed by the execution of a function are not constant, and depends on the particular input to that function}.
Similarly, if the cost incurred is power consumption, this may fluctuate randomly based on {the amount of instructions to be executed, the amount of input-output operations, etc. All these events are not constant when executing a function, since they depend on the input of that function}. 

Furthermore, in our model, the distribution of costs incurred and resources consumed are unknown to the controller initially, and it must learn the correct placement online. 

{Observe that resource contraints need to be satisfied on average. This means that ``overbooking'' is tolerated in some timeslot. Overbooking has been considered in cloud computing~\cite{chakraborty2024optimizing,overbooking} and is based on the idea that, even if the requirements \emph{declared} by the function owners are not met, functions can still work properly, as the actual amount of resources \emph{used} is generally less than the required ones. In the numerical results we will however show that in practice node capacities are not exceeded.}

\subsection{Optimal static policy}
\label{static}
Let us first consider an easier problem, where an omniscient oracle knows {the means of costs and resource requirements, i.e.,} values of $c_{i,k}$, $a_{i,j,k}$
and $\lambda_{j}$ for all $i,j,k$, and acts according to a static policy denoted by $p$, where $p_{i,j} = \mathbb{P}(i(t) = i| j(t) = j)$ is the probability of placing a function of type $j$ at node $i$.
{In every time slot~$t$ in which~$j~=~j(t)$, such a static policy would take decision~$I_{i,j}(t)$ so that~$\mathbb E[I_{i,j}(t)]~=~p_{i,j}, \forall i\in~[N]$.}
In this setting, the optimal static policy $p^\star$ is the optimal solution to the following linear program: 
\begin{align}
\label{eq:orig-optim}
\text{ min }_p & \sum_{i=1}^{N} \sum_{j=1}^{M}  p_{i,j} \lambda_{j} c_{i,j}   \\
\label{eq:orig-constr1}
\text{ s.t. }& \sum_{j=1}^{M} \lambda_{j} p_{i,j} { a_{i,j,k} \over \beta_{i,k} } \le 1 \quad\forall i,k\\ 
\label{eq:orig-constr2}
 \text{ with } & p\in \Delta^{M} = \{p \in [0,1]^{N \times M}: \sum_{i=1}^{N} p_{i,j} = 1 \quad \forall j\}
\end{align}
where $\Delta^{M}$ is the set of possible strategies. The oracle can compute $p^\star$ using an efficient algorithm for linear programming, for instance, interior point methods. To simplify notation, we define 

\begin{align*}
&\kappa_{i,j,k} = \lambda_{j} a_{i,j,k}/\beta_{i,k}, && \forall i\in[N],j\in[M],k\in[K]
\\
&f(p,c,\lambda) = \sum_{i=1}^N \sum_{j=1}^M p_{i,j} \lambda_j c_{i,j}
\\
&g_{i,k}(p,\kappa) = \sum_{j=1}^M p_{i,j} \kappa_{i,j,k} && \forall i\in[N], k\in[K]
\end{align*}

Therefore, problem~\eqref{eq:orig-optim}-\eqref{eq:orig-constr2} reduces to
\begin{align*}
&\text{ min } f(p,c,\lambda) \\
&\text{ s.t. } \max_{i,k} g_{i,k}(p,\kappa) \le 1 \text{ , }  p \in \Delta^{M}
\end{align*}

\subsection{Regret minimization}
\label{regret}
In the original problem \eqref{eq:resource-object}-\eqref{eq:resource-constraint_1}, the controller cannot perform as well as the oracle (Sec.~\ref{static}) because he has strictly less information. Moreover, problem~\eqref{eq:resource-object}-\eqref{eq:resource-constraint_1} is not rigorously defined, since it contains random variables. Denote by $\hat{p}(t)$ the policy used by the learner at time $t$, so that ${\hat p}_{i,j}(t) = \mathbb{P}(i(t) = i | j(t) = j)$. Therefore, we propose to minimize the expected regret~\cite{slivkins2019introduction}:
\begin{equation}
\label{regret_equ}
    R(T) = \sum_{t=1}^T  \mathbb{E}(f({\hat p}(t),c,\lambda) - f(p^\star,c,\lambda))
\end{equation}
while respecting the constraints with high probability, that is
\begin{equation}
\label{contraint_equ}
 \mathbb{P}( \max_{i,k} g_{i,k}({\hat p}(t),\kappa) \le 1) = 1 - o(1) \text{ when } T \to \infty   
\end{equation}

The regret is the sum of differences between the {the costs collected by the} controller and the oracle, so regret minimization is roughly equivalent to minimizing the time necessary to learn the optimal policy $p^\star$. If ${1 \over T} R(T)$ vanishes when $T \to \infty$, then the cost policy $\hat{p}(t)$ must converge to the optimal one $p^\star$ and the rate at which it vanishes measures how quickly the controller learns $p^\star$. Regret minimization is not trivial, because one must carefully sample the available decisions (including sub-optimal ones) to balance exploration and exploitation and also ensure that constraints are respected with high probability. We will propose two algorithms for regret minimization in the following sections.

\subsection{Generalizing the model}
\label{generalized}
To take into account admission control, we consider that node $N$ is a ``fictitious" node with infinite resources $\beta_{N,k} = +\infty$, and maximal cost $c_{N,j} = 1$ for all $j,k$. Placing a function at node $N$ is equivalent to rejecting it in the actual system. This ensures that a feasible solution of \eqref{eq:orig-optim}-\eqref{eq:orig-constr2} always exists. Since $c_{N,j} = 1$, the optimal solution always verifies $p^\star_{N,j} = 0$ as long as there exists a feasible solution to the original (without node $N$) problem. 

%% file: algorithm.tex
\section{Proposed algorithms}\label{section:algorithm}
\subsection{Base algorithm} 
\begin{figure}[ht]
\small
\begin{mdframed}[innerleftmargin=3pt,innerrightmargin=3pt]

    For $t \in [T]$:

\hspace{0.5cm} 1. Compute the statistics for all $i,j$:
\begin{align*}
n_{i,j}(t) &=  \sum_{s \in [t-1]} {I_j(s)\cdot} I_{i,j}(s) \\
n_{j}(t) &= \sum_{s\in[t-1]} I_j(s) =  \sum_{i=1}^N n_{i,j}(t) 
\end{align*}
\hspace{0.5cm} 2. Compute the estimates for all $i,j,k$:
\begin{align*}
{\hat \lambda}_{j}(t) &= n_{j}(t)/ (t-1)  \\
{\hat c}_{i,j}(t) &= \sum_{s \in [t-1]} C_{i,j}(s) I_{i,j}(s) /n_{i,j}(t) \\
{\hat \kappa}_{i,j,k}(t) &= \sum_{s \in [t-1]} (A_{i,j,k}(s) / \beta_{i,k})  I_j(s) /(t-1)
\end{align*}
\hspace{0.5cm} 3. Compute the confidence bounds for all $i,j,k$:
\begin{align*}
{\lambda}^-_{j}(t) &= \min \{ \lambda \in [0,1]: t D( {\hat \lambda}_{j}(t),\lambda) \le \ln t \}  \\
{c}^{-}_{i,j}(t) &= \min \{ c \in [0,1]: n_{i,j}(t) D( {\hat c}_{i,j}(t),c) \le \ln t \} \\
{\kappa}^{+}_{i,j,k}(t) &= \max \{ \kappa \in [0,1]: t D( {\hat \kappa}_{i,j,k}(t),\kappa) \le \ln t \} \text{ where } \\
D(\mu,\gamma) &= \mu \ln ( \mu  / \gamma) + (1-\mu)  \ln ( (1-\mu) / (1-\gamma) )
\end{align*}
\hspace{0.5cm} 4. Compute ${\hat p}(t)$ the minimizer of $f(p,c^{-}(t),\lambda^{-}(t))$ subject to $\max_{i \in [N], k \in [K]} g_{i,k}(p,\kappa^{+}(t)) \le 1$ and $p \in \Delta^{M}$.

\hspace{0.5cm} 5. Observe $i(t)$ and draw action $j(t)$ so that  $\mathbb{P}[ j(t) = j| {\cal F}(t)] = {\hat p}_{i,j(t)}(t)$ where ${\cal F}(t)$ represents the information available at time $t$. 
\end{mdframed}
    \caption{\label{base_algorithm} Base Algorithm.}
\end{figure}
We present the pseudo-code of the base version of the proposed algorithm in Figure~\ref{base_algorithm}. Let us break down the algorithm step by step.
\begin{enumerate}
    \item The learner first computes $n_{i,j}(t)$ the number of times a function of class $j$ was assigned to node $i$ and $n_{j}(t)$ and the number of times a function of class $j$ entered the network.
    \item The learner then computes estimates for the unknown parameters: the arrival rates $\lambda$, the costs $c$, and the resource consumption $\kappa$.
    \item Now, instead of using those estimates, the learner computes \emph{confidence bounds} for the unknown parameters, constructed such that, with high probability $\mathbf {\lambda^{-}} \stackrel{\text{ew}}{\le} \mathbf \lambda$, $\mathbf {c^{-}} \stackrel{\text{ew}}{\le} \mathbf c$ and $\mathbf {\kappa^{+}} \stackrel{\text{ew}}{\ge} \mathbf \kappa$. We use confidence bounds to force the algorithm to select sub-optimal actions sufficiently many times, as simply using estimates would not ensure enough exploration, resulting in high regret in some cases. The numbers ${\lambda}_j^-(t)$, ${c}^{-}_{i,j}(t)$ and ${\kappa}^{+}_{i,j,k}(t)$ are a lower confidence bound for $\lambda_j$, a lower confidence bound for $c_{i,j}$ and an upper confidence bound for $\kappa_{i,j,k}$ respectively. $D(\mu,\gamma)$ is the relative entropy between two Bernoulli distributions with means $\mu$ and $\gamma$, respectively. This strategy for constructing confidence intervals originates from the KL-UCB algorithm proposed by~\cite{kl_ucb}. Computing ${\lambda}^-_{j}(t)$ is equivalent to finding the unique zero of the function $\lambda \mapsto t D( {\hat \lambda}_{j}(t),\lambda) - \ln t$ over interval $[0,{\hat \lambda}_{j}(t)]$ which is both decreasing and differentiable, so one can do this computation very quickly using Newton's method. The same remark applies for ${c}^{-}_{i,j}(t)$ and ${\kappa}^{+}_{i,j,k}(t)$.
    \item the learner computes the chosen policy ${\hat p}(t)$ as the solution to the original optimization problem by replacing the unknown $\lambda$, $c$, $\kappa$ by the confidence bounds $\lambda^{-}$, $c^{-}$, $\kappa^{+}$.
    \item Using upper confidence bound $\kappa^{+}$ instead of $\hat{\kappa}$ forces the algorithm to be conservative since it implies the assumption that resource consumption would be at the higher end of the confidence interval. And using lower confidence bound for ${c}^{-}$ and $\lambda^{-}$ forces the algorithm to try to minimize the cost which leads to minimize the regret. In this way, the chosen policy will respect the constraints \eqref{contraint_equ} with high probability and choose decisions with low regret \eqref{regret_equ} when enough exploration has occurred. Finally, the learner draws the action according to the chosen policy ${\hat p}(t)$
\end{enumerate}
The optimization problem solved in Step 4 is a linear program with $NM$ variables and $N(1+M+K)$ constraints so that it can be solved in polynomial time by an efficient algorithm, for instance, an interior point method, and running $T$ iterations of this algorithm can be done in time $O(T( NMK + P(N,M,K)))$ and memory $O(NMK)$  where $P(N,M,K)$ is the time required to solve the optimization problem of Step 4, and is a polynomial in $N,M,K$. Since complexity is polynomial, we can apply the proposed algorithm to large problem instances.
\subsection{Fast algorithm}
We now propose a simple strategy to accelerate the base algorithm. Since the computed policy ${\hat p}(t)$ does not rapidly change between successive iterations, particularly when $t$ is large and ${\hat p}(t)$ is close to the optimal policy $p^\star$, we propose to perform Steps 1 to 4 infrequently. More precisely, given a parameter $\rho > 1$, define the update times ${\cal T} = ( \lceil \rho^{k} \rceil )_{k \in \mathbb{N}}$. If $t \in {\cal T}$, the learner performs Steps 1 to 5 as in the base algorithm, and if $t \not\in {\cal T}$, the learner skips Steps 1 to 4 and selects the strategy used at the previous time step $\hat{p}(t-1)$. Since the computationally heavy steps are only performed $| [T] \cup {\cal T}|$ times instead of $T$ times,  the speedup compared to the base algorithm equals roughly $T / |[T] \cup {\cal T}| = T / \lceil (\ln T)/(\ln \rho) \rceil$. As an illustration, for $\rho=1.05$ and $T=10^4$, the speedup factor is at least an order of magnitude, which is considerable. As shown in the numerical experiments, if we chose $\rho$ appropriately, the fast algorithm performs almost as well as the base algorithm while being an order of magnitude faster, enabling us to deal with potentially even larger problems. 

%% file: analysis.tex
\section{Algorithm Analysis}\label{section:analysis}
\subsection{Theoretical Performance Guarantees}
In this section we provide theoretical performance guarantees for the proposed algorithm which are summarized in Theorem~\ref{th:regret}. 
\begin{theorem}\label{th:regret}
Under the proposed algorithm the constraints are satisfied with high probability and the regret per unit of time vanishes at the rate:
\begin{align*}
{R(T)  \over T} &\le {N M \over T} [ 2( \sqrt{2} + 1) \sqrt{T \ln T} +  6 K (1 + \ln T) ] \\
&\underset{T \to \infty}{\sim} O\left( N M  \sqrt{\ln T \over T } \right) \\ 
\mathbb{P}& (\max_{i,k} g_{i,k}({\hat p}(t)) \le 1) \ge 1 - 6 N M K {\ln T \over T}\underset{T \to \infty}{\to} 1
\end{align*}
\end{theorem}
Theorem~\ref{th:regret} shows that the proposed algorithm both ensures that, asymptotically, the cost of the action distribution selected by the learner ${\hat p}(t)$ converges to the cost of the optimal action distribution $p^\star$. Furthermore, ${\hat p}(t)$ respects the constraints with high probability. In short, the proposed algorithm learns to act optimally, and its rate of convergence can be precisely evaluated. The proof is involved and presented in next subsection. 
\subsection{Proof of Theorem~\ref{th:regret}}
We first introduce following events
\begin{align*}
\mathfrak{C}(t) &= \cup_{i=1}^N \cup_{j=1}^M  \{ c_{i,j}^{-}(t) \not\in [c_{i,j} - \sqrt{ (2\ln t)/n_{i,j}(t)} , c_{i,j}] \} \\
\mathfrak{L}(t) &= \cup_{j=1}^M  \{ \lambda_j^{-}(t) \not\in [\lambda_j - \sqrt{(\ln t)/t]},\lambda_j] \}
\end{align*}
\begin{align*}
\small
\mathfrak{K}(t) &= \cup_{i=1}^N \cup_{j=1}^M \cup_{k=1}^K \{ \kappa^{+}_{i,j,k}(t) \not\in [\kappa_{i,j,k}, \kappa_{i,j,k} +  \sqrt{(\ln t)/t}]\}
\end{align*}
and their union $\mathfrak{U}(t)= \mathfrak{C}(t) \cup \mathfrak{L}(t) \cup \mathfrak{K}(t)$. Such a union represents the event that at least one among bounds $c_{i,j}^{-}(t)$,  $\lambda_{j}^{-}(t)$ and $\kappa_{i,j,k}^{+}(t)$ is ``too far'' from the respective true value. We will show that all those events occur with very small probability.

From Pinsker's inequality~\cite{thomas2006elements}, $D(\mu,\gamma) \ge 2(\mu-\gamma)^2$ so
\begin{align*}
c_{i,j}^{-}(t) &\ge {\hat c}_{i,j}(t) -  \sqrt{(\ln t)/(2 n_{i,j}(t))} \quad \forall i,j \\
\lambda_{j}^{-}(t) &\ge {\hat \lambda}_{j}(t) -  \sqrt{(\ln t)/(2t)}  \quad \forall j \\
\kappa_{i,j,k}^{+}(t) &\le {\hat \kappa}_{i,j,k}(t) +  \sqrt{(\ln t)/(2t)}  \quad \forall i,j,k
\end{align*}
Furthermore, from Hoeffding's inequality~\cite{hoeffding1994probability}, $\forall i,j,k$
\begin{align*}
&\mathbb{P}( {\hat \lambda}_j(t) \le \lambda_{j} - \sqrt{(\ln t)/(2t)} ) \le 1/t \\
&\mathbb{P}( {\hat \kappa}_{i,j,k}(t) \le \kappa_{i,j,k}- \sqrt{(\ln t)/(2t)} ) \le 1/t \\
&\mathbb{P}( {\hat c}_{i,j}(t) \le c_{i,j} - \sqrt{(\ln t)/n_{i,j}(t)} ) \\
&\le \sum_{s=1}^t \mathbb{P}( {\hat c}_{i,j}(t) \le c_{i,j} - \sqrt{(\ln t)/s},n_{i,j}(t) = s ) \\
&\le \sum_{s=1}^t (1/t^2) = 1/t 
\end{align*}
Finally from Garivier's inequality~\cite{kl_ucb}, we have
\begin{align*}
    \mathbb{P}( c_{i,j}^{-}(t) \ge c_{i,j}) &\le 1/t, \ \ \forall i,j,k\\
    \mathbb{P}( \lambda_{j}^{-}(t) \ge \lambda_{j}) &\le 1/t, \ \ \forall i,j,k \\
    \mathbb{P}( \kappa_{i,j,k}^{+}(t) \ge \kappa_{i,j,k}) &\le 1/t, \ \ \forall i,j,k 
\end{align*}
Taking a union bound over $i,j,k$ we have proven that 
\begin{align*}
    \mathbb{P}( \mathfrak{C}(t)) &\le 2MN/t \\
    \mathbb{P}( \mathfrak{L}(t)) &\le 2 M/t \\
    \mathbb{P}( \mathfrak{K}(t)) &\le 2 N M K /t
\end{align*}
And summing the above we get 
\begin{align*}
    \mathbb{P}( \mathfrak{U}(t)) \le 6 N M K /t
\end{align*}
So that indeed $\mathfrak{U}(t)$ is an atypical event. When $\mathfrak{U}(t)$ occurs, the regret is bounded by
\begin{align*}
& \sum_{t=1}^{T}  \mathbb{E}\left( \mathbbm{1} \{ \mathfrak{U}(t) \} ( f(\hat{p}(t), c, \lambda) - f(p^{\star},c,\lambda) ) \right) \\
&\le  \sum_{t=1}^{T}  \mathbb{P}(\mathfrak{U}(t)) \le  \sum_{t=1}^{T}  6 N M K /t \\
&\le 6 N M K (1 + \ln T)
\end{align*} 
using the fact that $f(p,c,\lambda) \le 1$ for any $p$ and a series integral comparison. 

If $\mathfrak{U}(t)$ does not occur, we have $\kappa_{i,j,k} \le \kappa^{+}_{i,j,k}(t) $, $\forall i,j,k$, so 
\begin{align*}
    g_{i,k}({\hat p}(t), \kappa) \le g_{i,k}({\hat p}(t), \kappa^{+}(t)) \le 1, \forall i,k
\end{align*}
In turn, since $\kappa^{+}_{i,j,k}(t) \le \kappa_{i,j,k} + \sqrt{ (\ln t)/t}$, $\forall i,j,k$, we have
\begin{align*}
    g_{i,k}(p^\star,\kappa^{+}(t)) \le 1+ M \sqrt{(\ln t)/t}, \forall i,k
\end{align*}
since $g_{i,k}(p^\star,\kappa) \le 1$. Now define $\tilde{p}(t)$ and $\tilde{p}_{N,j}(t)$ as
\begin{align*}
    \tilde{p}_{i,j}(t) &= \max( p^\star_{i,j} - M \sqrt{(\ln t)/t},0), \forall i \ne N \\
    \tilde{p}_{N,j}(t) &= 1 - \sum_{i < N} \tilde{p}_{N,j}(t)    
\end{align*}
One may readily check that $\tilde{p}(t)$ is constructed such that $g_{i,k}(\tilde{p}(t),\kappa^{+}(t)) \le 1$, $\forall i,k$, and that $f(\tilde{p}(t),c) \le  f(p^\star,c) + N M \sqrt{(\ln t)/t}$.

We have
\begin{align*}
& f(\hat{p}(t), c, \lambda) - \sum_{i=1}^{N} \sum_{j=1}^{M}  \lambda_{j} {\hat p}_{i,j}(t) \sqrt{(2\ln t)/n_{i,j}(t)} \\
&\le f(\hat{p}(t),c^{-}(t),\lambda^{-}(t)) \le f(\tilde{p}(t),c^{-}(t),\lambda^{-}(t))  \\
&\le f(\tilde{p}(t),c,\lambda) \le f(p^{\star},c,\lambda) + N M \sqrt{(\ln t)/t}
\end{align*}
where we used the fact that $\mathfrak{U}(t)$ does not occur  and the definition of ${\hat p}(t)$.

When $\mathfrak{U}(t)$ does not occur, the regret is bounded by
\begin{align*}
& \sum_{t=1}^{T}  \mathbb{E}( \mathbbm{1} \{ \bar{\mathfrak{U(t)}} \} ( f(\hat{p}(t), c, \lambda) - f(p^{\star},c,\lambda) ) ) \\
&\le \sum_{t=1}^{T} N M \sqrt{\frac{\ln t}{t}} + \sum_{t=1}^{T} \sum_{i=1}^{N} \sum_{j=1}^{M} \mathbb{E}\left( \lambda_{j} {\hat p}_{i,j}(t) \sqrt{\frac{2 \ln t}{n_{i,j}(t)}} \right)
\end{align*}
Since $\mathbb{E}(  I_{i,j}(t) | n_{i,j}(t)) = \lambda_j {\hat p}_{i,j}(t)$:
\begin{align*}
&\sum_{t=1}^{T} \mathbb{E}\left( \lambda_{j} {\hat p}_{i,j}(t) \sqrt{\frac{2 \ln t}{n_{i,j}(t)}} \right) =\sum_{t=1}^{T} \mathbb{E}\left( I_{i,j}(t) \sqrt{\frac{2 \ln t}{n_{i,j}(t)}} \right) \\
&\le \mathbb{E}( \sum_{\ell=1}^T  \sqrt{(2 \ln T)/\ell} |\{t \le T: I_{i,j}(t) = 1, n_{i,j}(t)=\ell  \})\\
&\le\sum_{\ell=1}^T  \sqrt{(2 \ln T)/\ell} \le \sqrt{ 2 \ln T} \int_{1}^{T} \sqrt{1 \over u} du \le  \sqrt{8 T \ln T}
\end{align*}
using the fact that $\ln t \le \ln T$, and $n_{i,j}(t) \in [T]$ is incremented whenever $I_{i,j}(t) = 1$. Similarly we also have
\begin{align*}
\sum_{t=1}^{T} N M \sqrt{\frac{\ln t}{t}} \le MN \sqrt{4 T \ln T}
\end{align*}

Summing over $i,j$ we get the bound
\begin{align*}
&\sum_{t=1}^{T}  \mathbb{E}( \mathbbm{1} \{ \bar{\mathfrak{U}(t)} \} ( f(\hat{p}(t), c, \lambda) - f(p^{\star},c,\lambda) ) ) \\
&\le 2( \sqrt{2} + 1) M N \sqrt{T \ln T}
\end{align*}
and putting everything together proves  the announced result:
\begin{align*}
R(T) &\le \sum_{t=1}^{T}  \mathbb{E}( \mathbbm{1} \{ {\mathfrak{U}(t)} \} ( f(\hat{p}(t), c, \lambda) - f(p^{\star},c,\lambda) ) )  \\
&+ \sum_{t=1}^{T}  \mathbb{E}( \mathbbm{1} \{ \bar{\mathfrak{U}(t)} \} ( f(\hat{p}(t), c, \lambda) - f(p^{\star},c,\lambda) ) ) \\
&\le {N M} [ 2( \sqrt{2} + 1) \sqrt{T \ln T} +  6 K (1 + \ln T) ]
\end{align*}


%% file: numerical_results.tex
\section{Numerical results}\label{section:numerical}
In this section we assess the practical performance of the proposed algorithm using numerical experiments. Our experiments are fully reproducible and the code for all the experiments is publicly available at \url{https://github.com/Free-Wei/Dynamic_virtual_network_placement}. The experiments are run on an ordinary laptop (Dell Inc. Precision 5760 with 11th Gen Intel® Core™ i7-11850H) using Python, no GPU acceleration or specialized equipment was used. 

All performance measures are computed over $50$ independent runs, and we display both the averages, as well as the maximal and minimal values across those runs (represented as shaded regions) which highlights the best and worst performance across those runs (the maximal value may indicate the worst performance in some criteria). We consider both versions of the proposed algorithm: the base algorithm and the fast algorithm.

We consider $A_{i,j,k}(t)$ and $C_{i,j}(t)$ i.i.d. Bernoulli variables with respective means $a_{i,j,k}$ and $c_{i,j}$ both in $[0,1]$. We consider uniform arrivals that is $\lambda_j = 1/M$ for all $j$.  We average over several random problem instances, generated by drawing both $a_{i,j,k}$ and $c_{i,j}$ uniformly at random in $[0,1]$ for all $i,j,k$ and setting $\beta_{i,k} = 0.1$ for all $i,k$ to consider a scenario where node resources are not abundant and not all functions can be placed at the same node. Note that we set that arrival rate for functions as 100 functions per second.

To further enhance the efficacy of the algorithm, we add a very slight amount of forced exploration, so that, by replacing $\hat{p}(t)$ by $\hat{p}'(t)$ defined as
\begin{equation*}
    \hat{p}_{i,j}'(t) = 
    \begin{cases}
    \frac{\hat{p}_{i,j}'(t)}{\sum_{i=1}^N \hat{p}_{i,j}(t)},&
    \text{if } \hat{p}_{i,j}'(t) \geq 10^{-3}\\
    \frac{\epsilon(t)}{\sum_{i=1}^N \hat{p}_{i,j}(t)}, & 
    \text{if } \hat{p}_{i,j}'(t) < 10^{-3} 
    \end{cases}
    \label{eq:en_link}
\end{equation*}
with $\epsilon(t) = 0.01(1-\frac{t}{T})$. This encourages to explore at least a little bit every node, and makes regret even smaller.

\subsection{Regret and feasibility}
\label{regret_num}



\begin{figure}[H]
\centering
\includegraphics[width=0.8\figsize]{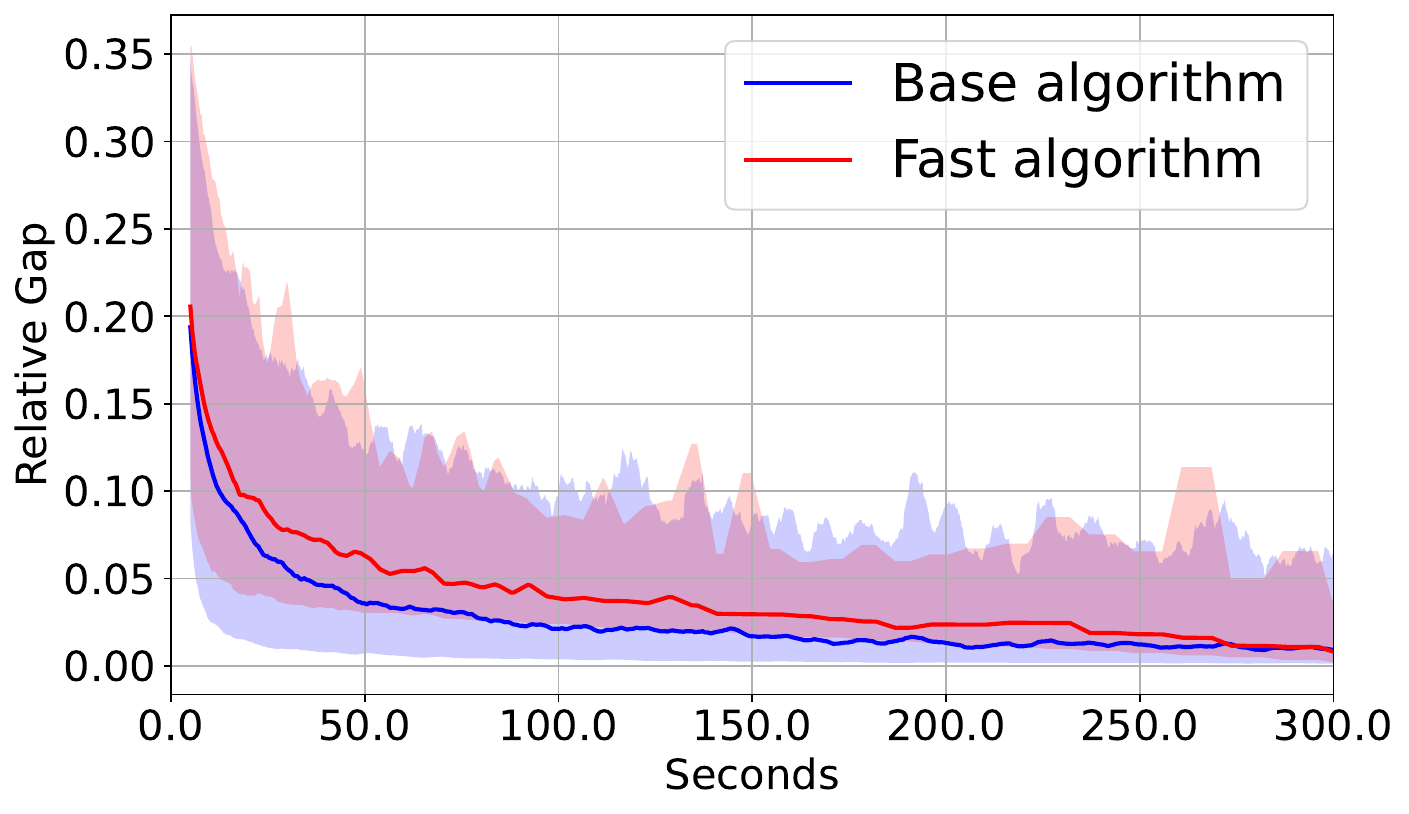}
\caption{Relative Gap under different algorithms.}\label{algo_diff}
\end{figure}

\begin{figure}[H]
\centering
\includegraphics[width=0.8\figsize]{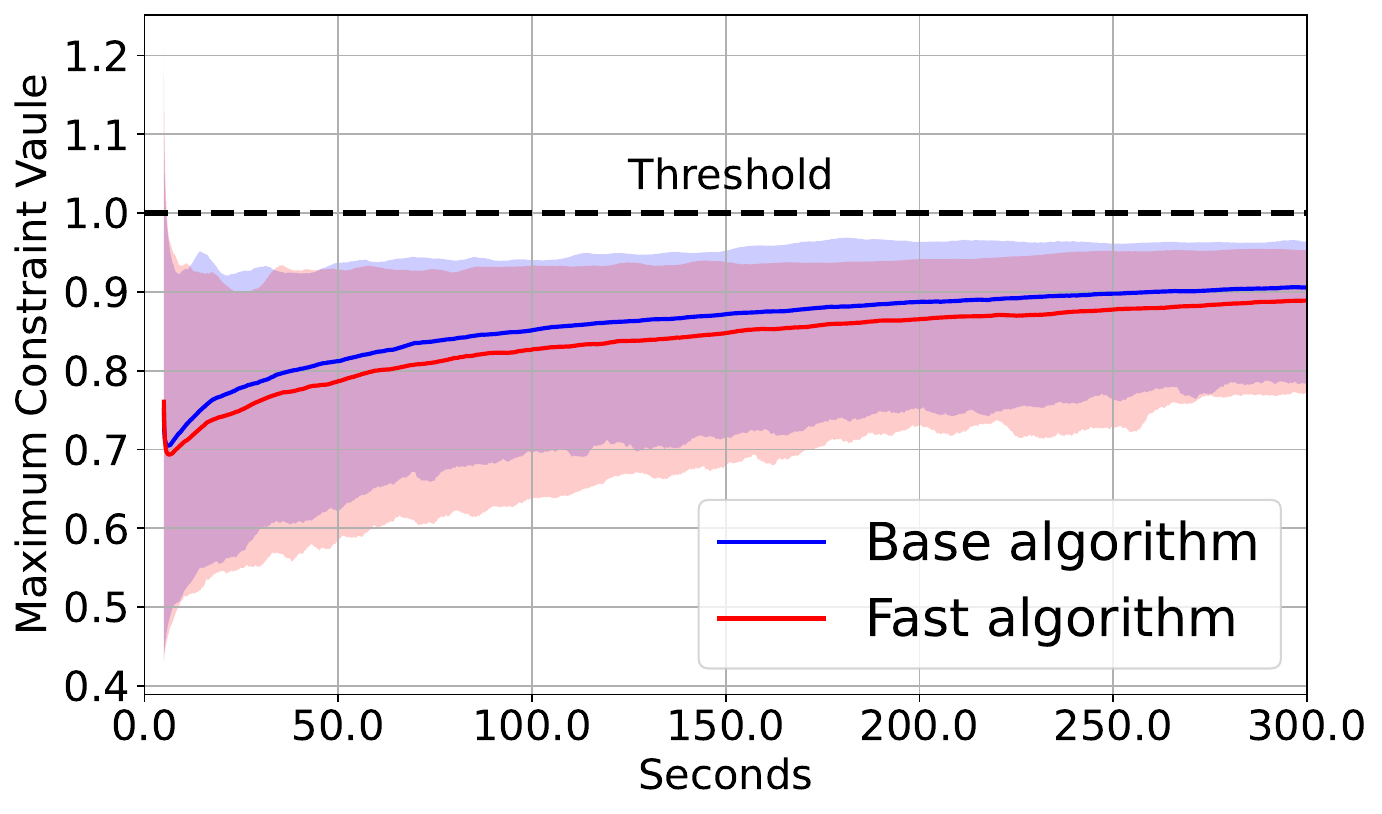}
\caption{Constraint value under different algorithms.}\label{algo_diff_cons}
\end{figure}

We first consider a relatively small scenario with $N=10$ nodes $M=3$ classes of functions and $K=2$ types of resources. In Figure~\ref{algo_diff} we present the relative performance gap between our proposed algorithm and the optimal policy:  
$[ f( \hat{p}(t),\lambda,c)- f( p^\star,\lambda,c)]/ f( p^\star,\lambda,c)$ as a function of the time $t$. In Figure~\ref{algo_diff_cons} we present the value of the largest constraint $\max_{i,k} g_{i,k}(\hat{p}(t),\kappa)$. Both figures show, as predicted by our theoretical results,  that the base algorithm converges quickly to the optimum policy (after $10^4$ iterations, the average relative gap is below $5\%$), and that with high probability (i.e. on all runs) the constraints $\max_{i \in [N] ,k \in [K]} g_{i,k}(\hat{p}(t),\kappa) \le 1$ are satisfied. Furthermore, we see that the fast algorithm behaves almost the same, and the gap between them is relatively small (for the fast algorithm we consider $\rho=1.05$). 

\subsection{Timing and acceleration}
\begin{table}[H]
\centering
\scalebox{1.2}{
\begin{tabular}{|l|c|c|c|}
\hline
\centering
{Algorithm} & $\tau_\text{mean}$ &  $\tau_\text{min}$ & $\tau_\text{max}$\\ 
\hline
Base & $4.053$ ms & $3.950$ ms  &  $4.153$ ms    \\
\hline
Fast ($\rho=1.05$)& $0.203$ ms  &$0.200$ ms  &  $0.210$ ms \\
\hline
\end{tabular}
}
\caption{Running time of one time slot for computing updates under the proposed algorithms.}\label{execution_time}
\end{table}


It is also important to check whether the amount of time needed to compute decisions via our algorithm can keep up with real system dynamics. To evaluate this, we compute the average time needed to calculate a decision at every time slot. Let~$\tau$ denote this average time. We compute~$\tau$ across the 50 independent runs by the base version and fast version with parameter~$\rho=1.05$, and show in Table~\ref{execution_time} the minimum, maximum and average value of~$\tau$ observed over such runs. We see that the fast version is faster by at least an order of magnitude (about $20$ times faster). Acceleration provides the best of both worlds: performance very close to the base algorithm, as well as a $20 \times$ speedup.

We also notice that most of the approaches from prior work based on neural networks require several days of training on a computer equipped with several GPUs, the proposed approach can learn the optimal policy in a matter of minutes on a common laptop, and is better suited to cases in which either computational resources are scarce, or the network conditions change frequently, so that the learning process must be restarted often, in order to track the changes and adapt. 

\subsection{Performance vs problem size}
\begin{figure}[H]
\centering
\includegraphics[width=0.8\figsize]{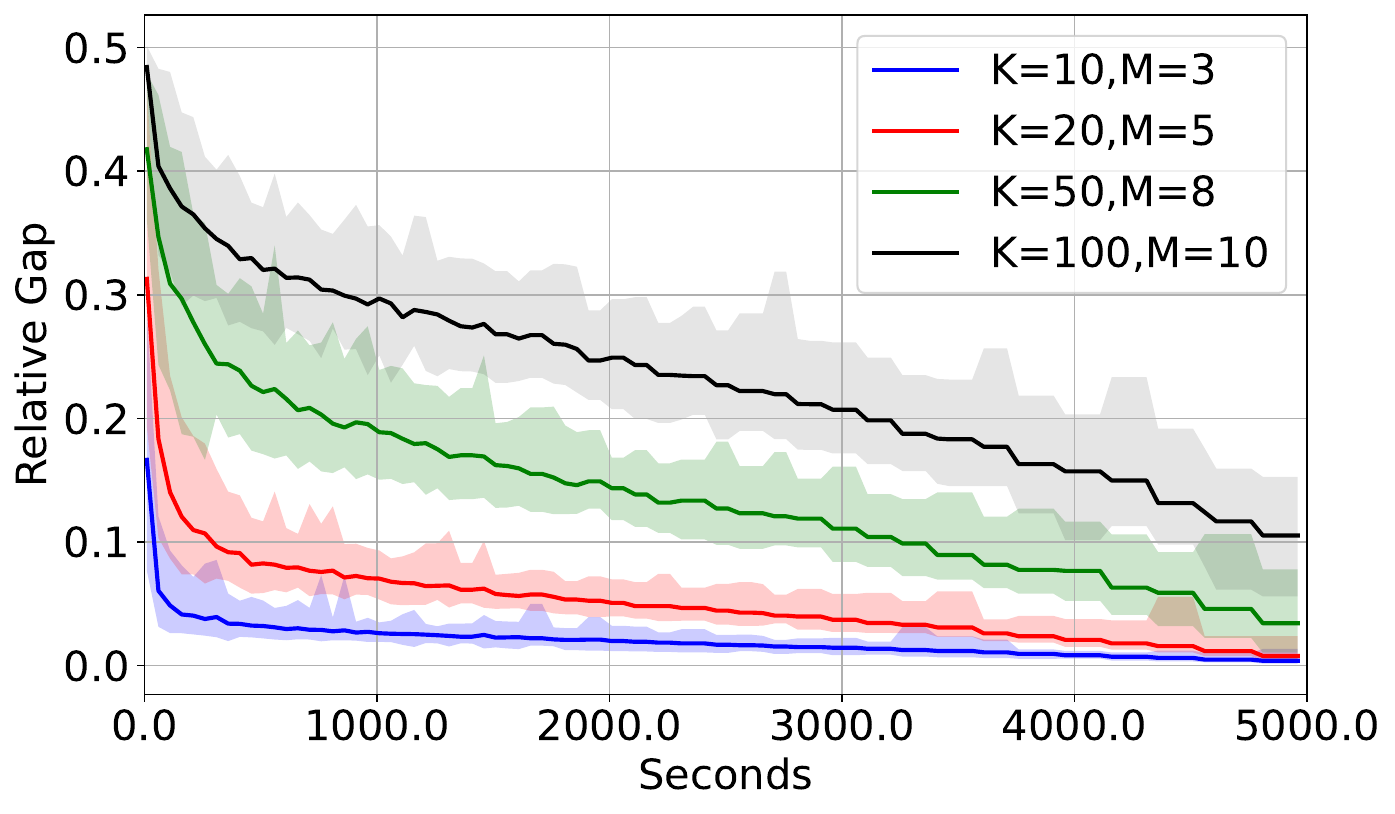}
\caption{Relative Gap under different scenarios.}\label{scenario_diff}
\end{figure}
\begin{figure}[H]
\centering
\includegraphics[width=0.8\figsize]{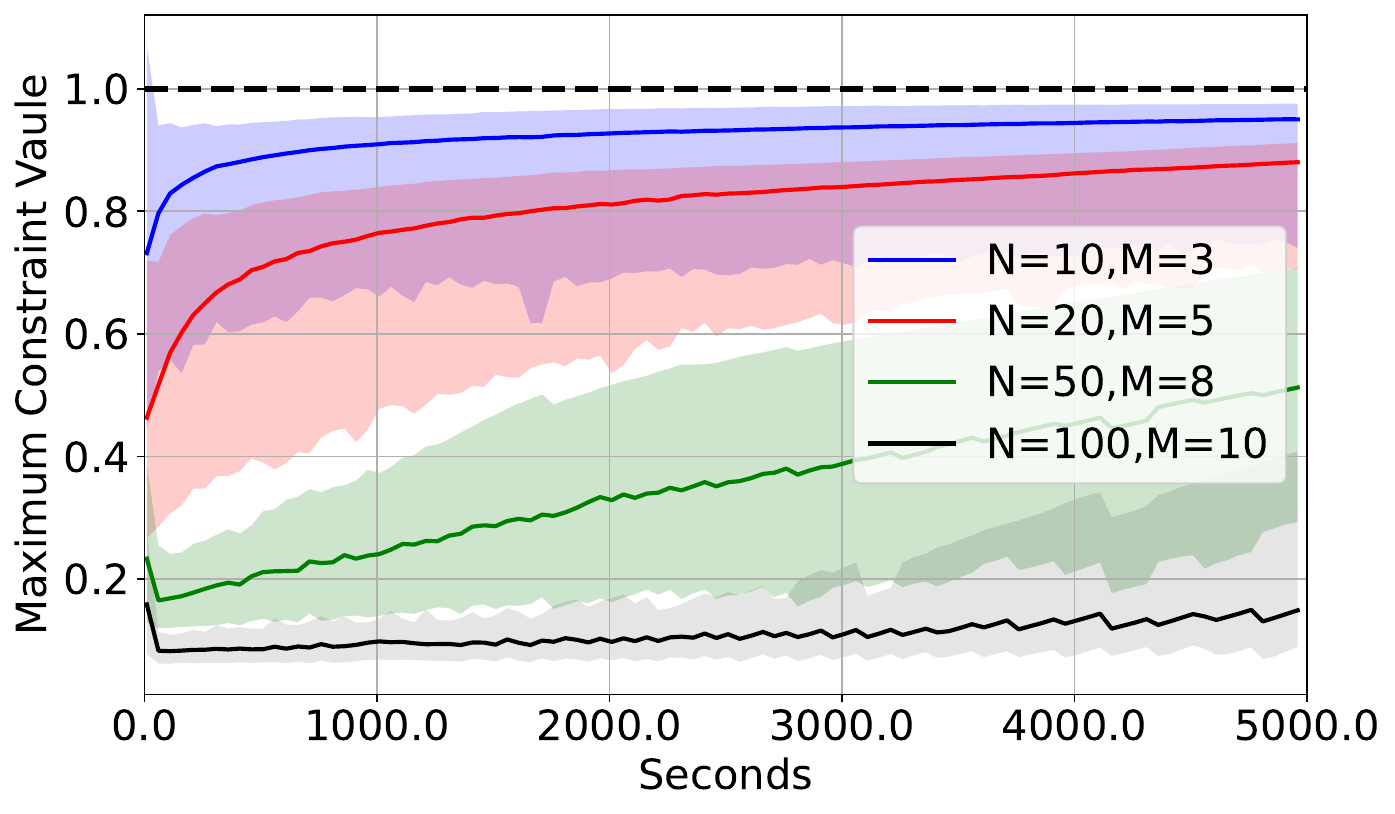}
\caption{Constraint value under different scenarios.}\label{scenario_diff_con}
\end{figure}

We now focus on how the performance scales as a function of the problem size that is $N \times M \times K$, to assess whether or not the good results obtained above can be generalized to large problem instances. We only display the performance of the fast version of the algorithm. In Figures~\ref{scenario_diff} and~\ref{scenario_diff_con} we present the relative performance gap and the value of the largest constraint respectively as a function of time for different values of $N$ and $M$. The number of iterations necessary to learn the optimal policy increases as a function of the product $N \times M$, however even for a relatively large problem with $N \times M = 10^3$ unknown parameters that must be estimated, the algorithm eventually learns the optimal policy quite well (less than $10\%$ relative gap) in $5 \times 10^3$ seconds. This is in line with our theoretical results which predict that the learning time should at most be proportional to $N^2 \times M^2$. 

\begin{table}
\centering
\scalebox{1.05}{
\begin{tabular}{|l|c|c|c|}
\hline
\centering
Problem Size & $\tau_\text{mean}$ &  $\tau_\text{min}$ & $\tau_\text{max}$\\ 
\hline
$(N,M,K) = (10,3,2)$ & 0.190 ms & 0.184 ms & 0.204 ms   \\
\hline
$(N,M,K) = (20,5,2)$ & 0.252 ms & 0.249 ms & 0.259 ms   \\
\hline
$(N,M,K) = (50,8,2)$ & 0.406 ms & 0.397 ms & 0.416 ms  \\
\hline
$(N,M,K) = (100,10,2)$ & 0.567 ms & 0.557 ms & 0.589 ms   \\
\hline
\end{tabular}
}
\caption{Running time of one time slot for computing updates under the fast algorithm vs problem size.}
\label{execution_time_NM}
\end{table}

In table \ref{execution_time_NM} we display the average, minimum and maximum time (in milliseconds) required to calculate a decision in one time slot by the fast version of the proposed algorithm, once again with parameter $\rho=1.05$ as a function of $M,N,K$ and we observe that, even with large problems the typical computation time for a run is below $5$ minutes, once again in stark contrast with neural network approaches which require several days of training for good performance.

\subsection{Fast version under different $\rho$}

\begin{figure}[h]
\centering
\includegraphics[width=0.8\figsize]{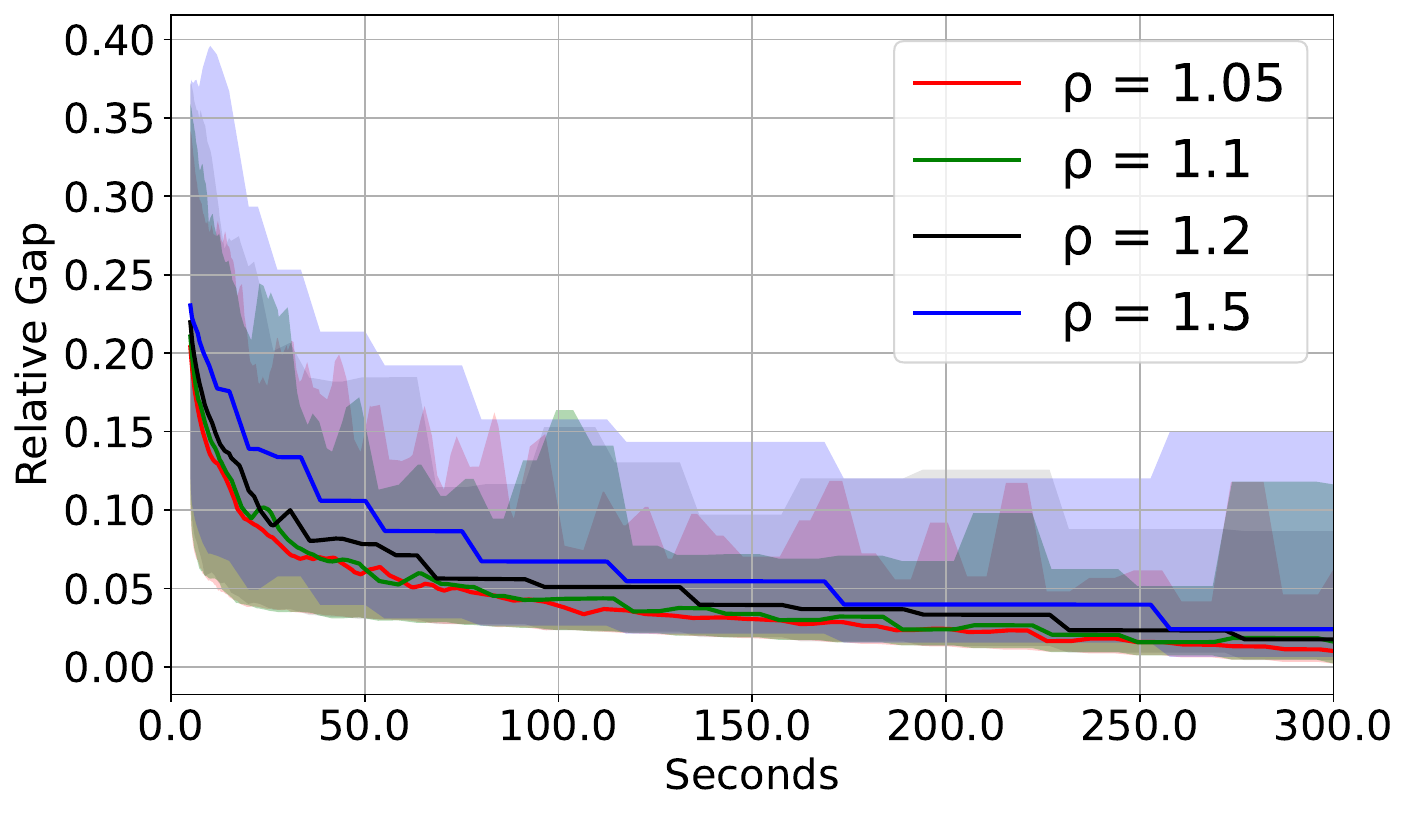}
\caption{Relative Gap under different $\rho$.}\label{scenario_diff_rho}
\end{figure}
\begin{figure}[H]
\centering
\includegraphics[width=0.8\figsize]{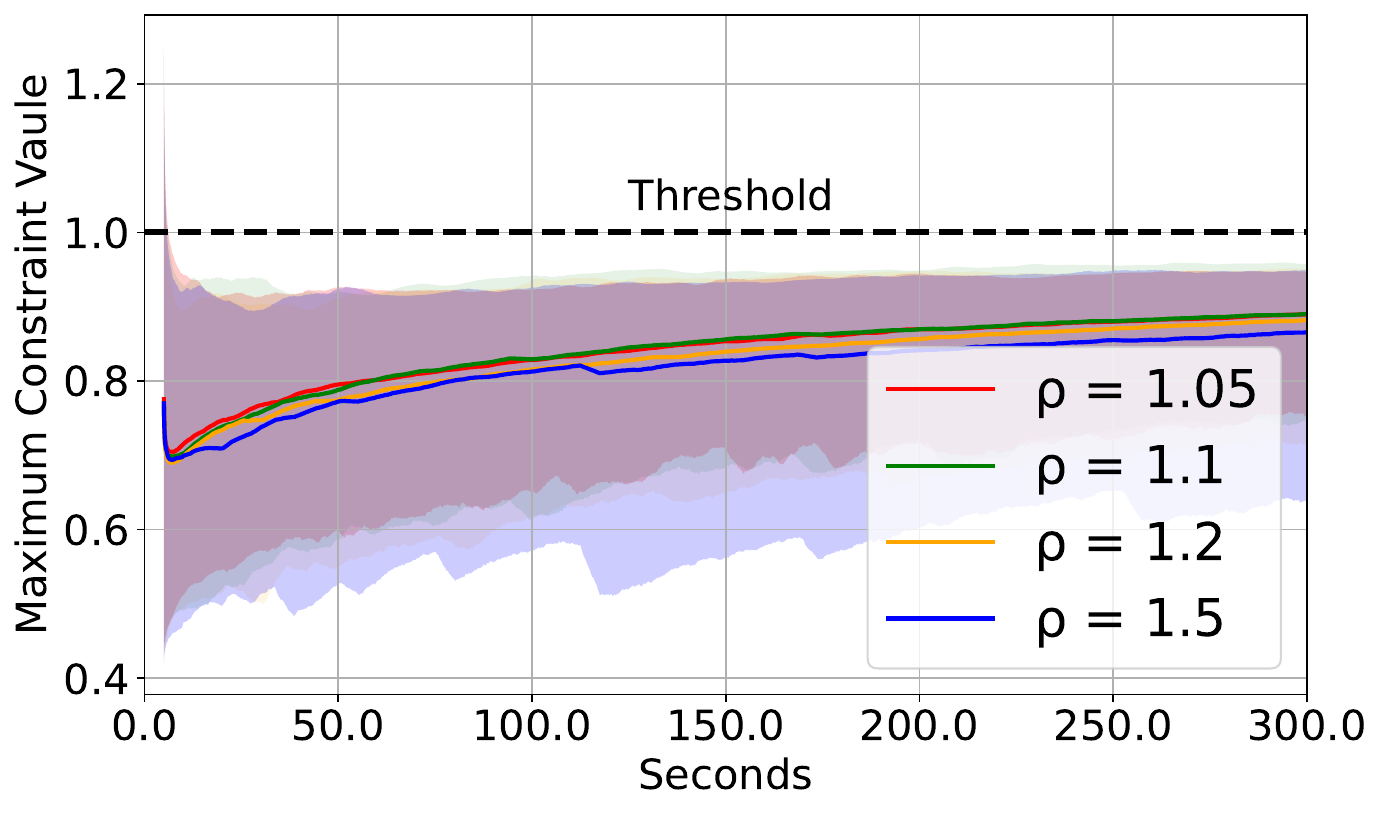}
\caption{Constraint value under different $\rho$.}\label{scenario_diff_con_rho}
\end{figure}

In the fast version, $\rho$ is selected as the update rate to let the controller skip some steps (e.g. computing all confidence bound, solving LP problem) to avoid useless updating, which leads to 20 $\times $speedup. However, if $\rho$ is too large, then necessary updates will also be missed. Thus, it is important to choose suitable $\rho$ such that saving running time while achieving optimal strategy. In Figure~\ref{scenario_diff_rho} and Figure~\ref{scenario_diff_con_rho}, we show relative gap and constraint value under different $\rho$ for the same setting in Section~\ref{regret_num}. It illustrates that $\rho = 1.05$ performs better than others and almost achieving the optimal policy at $300$ seconds. And we also notice that when $\rho = 1.5$, the confidence interval is big which means sometimes it may perform badly. In the Table~\ref{execution_time_rho}, we find the speedup is not huge when compared between $\rho = 1.05$ and $\rho = 1.5$. Therefore, considering their performances, $\rho = 1.05$ is better than $\rho = 1.5$. In conclusion, the trade-off between running time and accuracy may be well considered under different setting.

\begin{table}[H]
\centering
\scalebox{1.3}{
\begin{tabular}{|l|c|c|c|}
\hline
\centering
Different $\rho$  &$\tau_\text{mean}$ &  $\tau_\text{min}$ & $\tau_\text{max}$\\ 
\hline
$\rho =1.05$ & $0.218$ ms  & $0.202$ ms  & $0.241$ ms   \\
\hline
$\rho =1.1$ & $0.209$ ms & $0.193$  ms & $0.229$ ms   \\
\hline
$\rho =1.2$ & $0.203$  ms & $0.188$   ms &  $0.224$ ms  \\
\hline
$\rho =1.5$ & $0.199$ ms &  $0.185$  ms & $0.216$ ms   \\
\hline
\end{tabular}
}
\caption{Running time of one round for computing updates under different $\rho$.}
\label{execution_time_rho}
\end{table}

%% file: conclusion.tex
\section{Conclusion}\label{section:conclusion}
We propose a model for the virtual function placement problem and novel algorithms to learn the optimal placement policies based on multi-armed bandits. We show that their regret grows at a rate $O( N M \sqrt{T \ln T} )$ while respecting the feasibility constraints with high probability. Numerical experiments further show that they perform well in practice, with modest computational time, unlike neural network-based approaches. For future work, extending our results to the case where costs are correlated would be a non-trivial and interesting problem. 

%% file: acknowledgment.tex
\section{Acknowledgment}
This work was partially funded by Beyond5G, a project of the French Government’s recovery plan ``France Relance''.

%% file: main.bbl
\begin{thebibliography}{10}
\providecommand{\url}[1]{#1}
\csname url@samestyle\endcsname
\providecommand{\newblock}{\relax}
\providecommand{\bibinfo}[2]{#2}
\providecommand{\BIBentrySTDinterwordspacing}{\spaceskip=0pt\relax}
\providecommand{\BIBentryALTinterwordstretchfactor}{4}
\providecommand{\BIBentryALTinterwordspacing}{\spaceskip=\fontdimen2\font plus
\BIBentryALTinterwordstretchfactor\fontdimen3\font minus \fontdimen4\font\relax}
\providecommand{\BIBforeignlanguage}[2]{{%
\expandafter\ifx\csname l@#1\endcsname\relax
\typeout{** WARNING: IEEEtran.bst: No hyphenation pattern has been}%
\typeout{** loaded for the language `#1'. Using the pattern for}%
\typeout{** the default language instead.}%
\else
\language=\csname l@#1\endcsname
\fi
#2}}
\providecommand{\BIBdecl}{\relax}
\BIBdecl

\bibitem{sfc_placment_routing}
N.~He, S.~Yang, F.~Li, S.~Trajanovski, L.~Zhu, Y.~Wang, and X.~Fu, ``Leveraging deep reinforcement learning with attention mechanism for virtual network function placement and routing,'' \emph{IEEE Transactions on Parallel and Distributed Systems}, 2023.

\bibitem{Grosu2020}
H.~Badri, T.~Bahreini, D.~Grosu, and K.~Yang, ``Energy-aware application placement in mobile edge computing: A stochastic optimization approach,'' \emph{IEEE Transactions on Parallel and Distributed Systems}, 2020.

\bibitem{sfc_placement_tpds}
X.~Wang, H.~Xing, F.~Song, S.~Luo, P.~Dai, and B.~Zhao, ``On jointly optimizing partial offloading and sfc mapping: A cooperative dual-agent deep reinforcement learning approach,'' \emph{IEEE Transactions on Parallel and Distributed Systems}, 2023.

\bibitem{Grosu2022}
T.~Bahreini, H.~Badri, and D.~Grosu, ``Mechanisms for resource allocation and pricing in mobile edge computing systems,'' \emph{IEEE Transactions on Parallel and Distributed Systems}, vol.~33, no.~3, pp. 667--682, 2022.

\bibitem{slivkins2019introduction}
A.~Slivkins \emph{et~al.}, ``Introduction to multi-armed bandits,'' \emph{Foundations and Trends{\textregistered} in Machine Learning}, vol.~12, no. 1-2, pp. 1--286, 2019.

\bibitem{kl_ucb}
A.~Garivier and O.~Capp{\'e}, ``The kl-ucb algorithm for bounded stochastic bandits and beyond,'' in \emph{COLT 2011}.\hskip 1em plus 0.5em minus 0.4em\relax JMLR Workshop and Conference Proceedings, 2011, pp. 359--376.

\bibitem{dehury2019}
C.~K. Dehury and P.~K. Sahoo, ``Dyvine: Fitness-based dynamic virtual network embedding in cloud computing,'' \emph{IEEE Journal on Selected Areas in Communications}, 2019.

\bibitem{faas_placement}
N.~Mahmoudi, C.~Lin, H.~Khazaei, and M.~Litoiu, ``Optimizing serverless computing: introducing an adaptive function placement algorithm,'' in \emph{CASCON '19}.\hskip 1em plus 0.5em minus 0.4em\relax USA: IBM Corp., 2019.

\bibitem{yan2020}
Z.~Yan, J.~Ge, Y.~Wu, L.~Li, and T.~Li, ``Automatic virtual network embedding: A deep reinforcement learning approach with graph convolutional networks,'' \emph{IEEE Journal on Selected Areas in Communications}, vol.~38, no.~6, pp. 1040--1057, 2020.

\bibitem{zhang2022}
P.~Zhang, C.~Wang, N.~Kumar, W.~Zhang, and L.~Liu, ``Dynamic virtual network embedding algorithm based on graph convolution neural network and reinforcement learning,'' \emph{IEEE Internet of Things Journal}, 2022.

\bibitem{yao2021}
H.~Yao, B.~Zhang, P.~Zhang, S.~Wu, C.~Jiang, and S.~Guo, ``Rdam: A reinforcement learning based dynamic attribute matrix representation for virtual network embedding,'' \emph{IEEE Transactions on Emerging Topics in Computing}, 2021.

\bibitem{fu2021}
P.~G.~T. Jing~Fu, Bill~Moran and C.~Xing, ``Resource competition in virtual network embedding,'' \emph{Stochastic Models}, vol.~37, no.~1, pp. 231--263, 2021.

\bibitem{elkael2022}
M.~Elkael, M.~{Ait Aba}, A.~Araldo, H.~Castel-Taleb, and B.~Jouaber, ``Monkey business: Reinforcement learning meets neighborhood search for virtual network embedding,'' \emph{Computer Networks}, vol. 216, pp. 1389--1286, 2022.

\bibitem{maleki2023}
V.~Maleki~Raee, A.~Ebrahimzadeh, R.~H. Glitho, M.~El~Barachi, and F.~Belqasmi, ``E2dne: Energy efficient dynamic network embedding in virtualized wireless sensor networks,'' \emph{IEEE trans. green commun. netw.}, 2023.

\bibitem{elkael2023joint}
M.~Elkael, A.~Araldo, S.~d'Oro, H.~Castel-Taleb, M.~A. Aba, and B.~Jouaber, ``Joint placement, routing and dimensioning at the network edge for energy minimization,'' in \emph{GLOBECOM 2023}.\hskip 1em plus 0.5em minus 0.4em\relax IEEE, 2023, pp. 941--946.

\bibitem{chakraborty2024optimizing}
T.~Chakraborty, C.~Kopp, and A.~N. Toosi, ``Optimizing renewable energy utilization in cloud data centers through dynamic overbooking: An mdp-based approach,'' \emph{IEEE Transactions on Cloud Computing}, 2024.

\bibitem{overbooking}
M.~Liwang and X.~Wang, ``Overbooking-empowered computing resource provisioning in cloud-aided mobile edge networks,'' \emph{IEEE/ACM Transactions on Networking}, vol.~30, no.~5, pp. 2289--2303, 2022.

\bibitem{thomas2006elements}
M.~Thomas and A.~T. Joy, \emph{Elements of information theory}.\hskip 1em plus 0.5em minus 0.4em\relax Wiley-Interscience, 2006.

\bibitem{hoeffding1994probability}
W.~Hoeffding, ``Probability inequalities for sums of bounded random variables,'' \emph{The collected works of Wassily Hoeffding}, pp. 409--426, 1994.

\end{thebibliography}
